# A Smart Handheld Edge Device for On-Site Diagnosis and Classification of Texture and Stiffness of Excised Colorectal Cancer Polyps

Ozdemir Can Kara*[1], Jiaqi Xue*[1], Nethra Venkatayogi[1], Tarunraj G. Mohanraj[1], Yuki Hirata[2], Naruhiko Ikoma[2], S. Farokh Atashzar[3], and Farshid Alambeigi[1]

*Abstract*— This paper proposes a smart handheld textural sensing medical device with complementary Machine Learning (ML) algorithms to enable on-site Colorectal Cancer (CRC) polyp diagnosis and pathology of excised tumors. The proposed unique handheld edge device benefits from a unique tactile sensing module and a dual-stage machine learning algorithms (composed of a dilated residual network and a t-SNE engine) for polyp type and stiffness characterization. Solely utilizing the occlusion-free, illumination-resilient textural images captured by the proposed tactile sensor, the framework is able to sensitively and reliably identify the type and stage of CRC polyps by classifying their texture and stiffness, respectively. Moreover, the proposed handheld medical edge device benefits from internet connectivity for enabling remote digital pathology (boosting the diagnosis in operating rooms and promoting accessibility and equity in medical diagnosis).

## I. INTRODUCTION

Colorectal cancer (CRC) is the second leading cause of cancer-related deaths worldwide [1]. New cases of colorectal cancer are estimated to exceed 3.2 million by 2040, based on projections of population growth and aging. On the other hand, 91% of the CRC patients can fully recover if early-stage cancer can be promptly diagnosed [2]. Thus, a reliable diagnosis of pre-cancerous lesions (e.g., polyps) with a convenient diagnostic framework becomes essential to reduce mortality and morbidity and increase the treatment alternatives [3].

Recent literature reports that morphological characteristics of polyps (i.e., shape, size, and particularly texture) can aid clinicians with early diagnosis and classifications of cancer types and stages, treatment outcomes, and, more importantly, survival of cancer patients [4]. Despite their impactful attributes, decision-making based on morphological characteristics can depend on the clinician and be heavily subjective, unreliable, and conflicting [5]. This has been mainly caused by the limitations of existing vision-based colonoscopic technologies and polyps classification approaches, creating a challenge for an early diagnosis of CRC. Besides textural features, it has also been revealed that the progress of cancer, such as cell invasion, transformation, migration as well as cancer stage, are highly associated with the change in the polyp's modulus of elasticity/stiffness [6]. In other words, when cancer shows pathological development and progress from an early stage to the later stages, clinicians mostly observe the increase in the modulus of elasticity of polyps [7]. This valuable piece of information can be considered a key feature if characterized by an accurate and reliable diagnostic framework.

Currently, for a complete and accurate diagnosis and treatment of CRC, typically, a biopsy sample or excised polyp is first collected during a colonoscopic or surgical procedure and sent to the pathology lab for the analysis of histology [8]. Next, to gather digitalized histology slides of the biopsied sample, laboratory technicians follow a manual and tedious procedure [9]. This procedure includes collection and fixation, dehydration and clearing, paraffin embedding, microtomy, staining, mounting, and digitalization of the

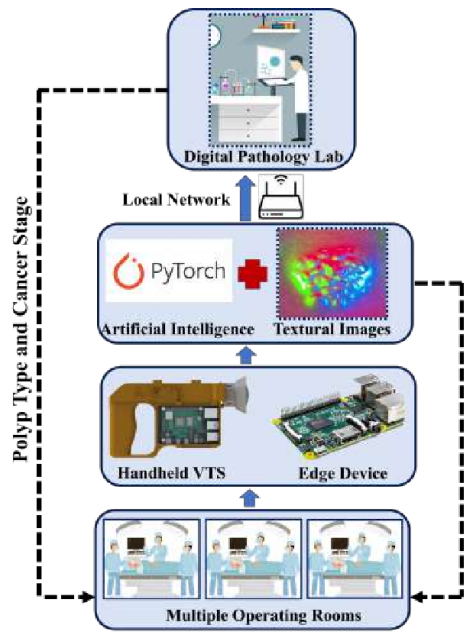

Fig. 1. Conceptual illustration of the proposed smart digital pathology in a hospital. Using the proposed device multiple operating rooms can be connected through local network and the edge device to a remote pathology lab.

Research reported in this publication was supported by the National Cancer Institute of the National Institutes of Health under Award Number R21CA280747.

*Authors contributed equally to this work.

[1]Ozdemir Can Kara, Nethra Venkatayogi, Tarunraj G. Mohanraj, Jiaqi Xue and Farshid Alambeigi are the with Texas Robotics at The University of Texas at Austin, Austin, TX, USA, 78712. Email: {ozdemirckara, venkatayoginethra, tarunrajgm, jiaqixue}@utexas.edu, and {farshid.alambeigi}@austin.utexas.edu.

[2]Yuki Hirata and Naruhiko Ikoma are with the Department of Surgical Oncology, Division of Surgery, The University of Texas MD Anderson Cancer Center, Houston, TX, USA, 77030. Email: {yhirata,nikoma}@mdanderson.org.

[3]S. Farokh Atashzar is with Department of Electrical and Computer Engineering, and Department of Mechanical and Aerospace Engineering, New York University, NY, USA, 11201. Email: {f.atashzar@nyu.edu}.

biopsied samples [10]. It is imperative that the biopsied sample should be handled to preserve its internal structure and display a similar appearance as inside the human body. Nevertheless, all of the aforementioned procedural steps can generate various artifacts that may reduce the quality of the histological images [11]. Aside from these issues, access time to a pathology lab to deliver biopsy samples and the decision-making process and time are the other critical challenges of traditional pathology [11]. These shortcomings, therefore, demand the development of new on-site technologies that can obtain textural and stiffness features of biopsied samples and excised CRC polyps inside the operating room. More importantly, they should enable providing this critical information to the clinicians available inside the room and also the pathologists who may work remotely in another location.

Vision-based Tactile Sensors (VTSs) have recently been developed to improve tactile perception through high-resolution visual information [12]. These sensors can provide high-resolution 3D visual image reconstruction and localization of the interacting objects by capturing tiny deformations of an elastic gel layer that directly interacts with the objects' surface [12], [13]. For instance, Johnson and Adelson [14] developed the GelSight sensor that has been used for various non-medical applications, including the surface texture recognition, geometry measurement with deep learning algorithms, and localization and manipulation of small objects [12]. Despite the great features of VTS, most of the existing examples require a relatively high interaction force between the sensor's gel layer which would not be acceptable in the clinical domain, particularly CRC diagnosis [15]. To address this critical shortcoming, Kara et al. [16] have recently proposed the first design of a new family of hyper-sensitive and high-fidelity VTS, called HySenSe. This new design can provide high-fidelity textural images of hard and even deformable objects with less than 1.5 N interaction force, shedding light on the uncovered potential of VTS in the medical domain for sensitive applications such as CRC polyp diagnosis and pathology.

Over the years, various computer-aided artificial intelligence (AI) algorithms and, recently, deep neural networks have been advancing in the field of CRC diagnosis and "digital pathology" to assist classification of the polyps mainly based on medical imaging [11], [17]–[19]. Despite their impact, the performances of artificial intelligence modules processing images such as colonoscopic scans are degraded by inherent frequent artifacts, including camera occlusion, suboptimal illumination, and inaccurate line of sight, affecting the quality of obtained images [20]. Moreover, most of the AI algorithms and deep networks require a large, annotated, and balanced dataset, which is not possible specifically considering the inter- and intra-patient variability [20]. Even though for CRC diagnosis, AI-driven digital pathology models have been used for polyp identification and classifications (mainly using visual information provided by an endoscopic camera), obtaining the classification of the polyp's type and recognizing the stiffness have not been achieved in the literature. In this study, we target this unmet need and propose the design of a handheld scalable VTS device to process textural information and conduct stiffness recognition and polyp classification.

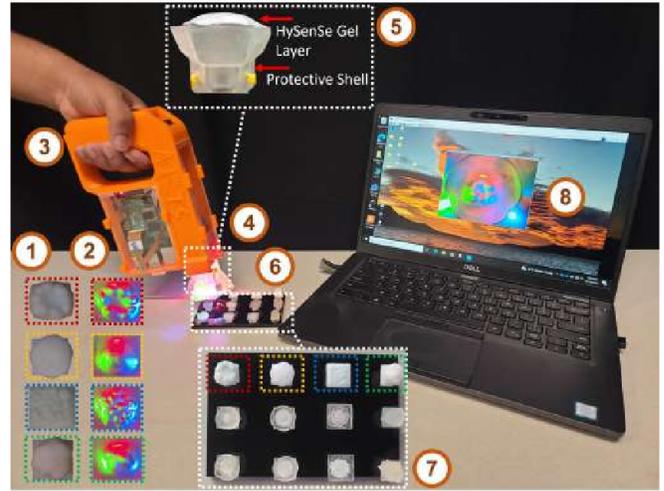

Fig. 2. The prototype of the proposed handheld edge device with a tactile sensing device towards digital pathology of the CRC polyps. Each polyp is fabricated based on realistic clinical polyps [21].①- Fabricated polyp phantom textures, ②- HySenSe visual outputs for each polyp phantom, ③- Proposed smart handheld edge device, ④- Sensor module including HySenSe, ⑤-Zoomed view of the sensor module, ⑥- Polyp phantom table including each material, ⑦- Zoomed view of the polyp phantom table, ⑧- HySense output from the sensor module camera in the remote laptop screen.

To collectively address the existing challenges in performing an on-site CRC polyp diagnosis and pathology of excised tumors, and as our main contributions, we propose a novel framework including a handheld edge device with a tactile sensing module and complementary AI algorithms. The handheld edge device (shown in Fig. 2) solely utilizes occlusion-free illumination-resilient textural images provided by a new modular and compact design of HySenSe sensor and is able to sensitively and reliably identify the type and stage of CRC polyps by classifying their type and stiffness. Moreover, as shown in Fig. 1, the implemented handheld medical edge benefits from internet connectivity and thus can be connected remotely to a remote pathology lab and readily transfer data to the pathologist who is not present in the operating room and address the aforementioned challenges of current digital pathology frameworks.

## II. METHODOLOGY

### A. Handheld Tactile Sensing Device

To satisfy the needs and constraints for performing a CRC polyp diagnosis and pathology inside the operating room, several requirements were taken into consideration during the design and development of the handheld tactile sensing device. Particularly, we were interested in designing an *untethered*, *modularized*, *locally/remotely controllable*, and *edge* device. The *untethered* design requirements were considered to satisfy the sterilization and hygiene operation of the device within an operating room and more importantly not change the surgical workflow. To address this need, an independent power supply was needed that could support the entire system running for more than 30 minutes of operation without requiring an external power cord. Moreover, the ability to locally and/or remotely (e.g., outside an operating room) control and monitor the diagnosis process provides a unique opportunity for real-time "collective diagnosis" using simultaneous collaboration between the local operator and the remote supervisor to perform the required tasks (i.e., screening, data recording, and real-time AI diagnosis) without noticeable interference (i.e., delay) to the process. As shown in Fig. 1, including an *edge device* (i.e., Raspberry Pi 4B (Raspberry Pi Foundation, UK)) within the structure of the handheld device ensures remote transfer of the obtained textural images to the pathology lab and collect the expert opinion of a remote pathologist who cannot be present in the operating room. This critical feature can promote access and equity to high-quality diagnosis, which is critical during the operation, for example, for decision-making on the margins of cut or type of intervention as the surgery progresses. Finally, the *modularized* structure increased the usability and sterilization of the device after each operation. To satisfy this design need, we decomposed the complete system into three detachable and fast-assembly/disassembly modules including the sensor module, the computer module, and the handle. Of note, these modules will be described in detail in the following.

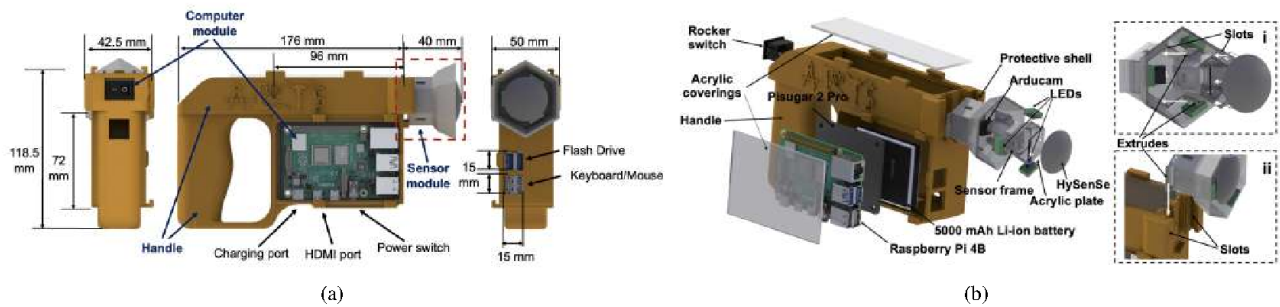

Fig. 3. Overall design of a proposed handheld textural tactile sensing device : (a) Assembled views, modules, and the dimension of the handheld device, (b) Exploded view of the proposed handheld device, (i) Interference connection between sensor module components, and (ii) Sliding mechanism for connecting the sensor module to the handle.

**Sensor module** includes a Hypersensitive Vision-Based Tactile Sensor (HySenSe), a sensor frame, and a protective shell as illustrated in Fig. 3(a). Of note, the key component of the HySenSe is a dome-shape soft silicone layer that can capture and characterize tiny textures of a CRC polyp during its safe examinations [22]. In this study, we first followed the procedure described in [16] to fabricate a HySense VTS including a gel layer with 00-20 Shore hardness (measured by Shore 00 scale durometer– Model 1600 Dial 169 Shore 00, Rex Gauge Company). The HySenSe was then glued and fixated to a transparent acrylic plate, while the acrylic plate, the sensor frame, and the protective shell were assembled together and secured to the device handle through slider mechanisms with interference as demonstrated in Fig. 3(b)[i][ii]. Three high-power RGB LEDs (941-XBDRED0801, 941-XBDGRN0D01, 941-XBDBLU0202) were fixed on the inner wall of the protective shell at 120 degrees apart. Of note, LEDs create color contrast and provide illumination for 3D textural details recreation during the interaction between the gel layer and the CRC polyps. The attachment of LEDs on the protective shell instead of the sensor frame provided a way to change the gel layer when needed, without interfering with the electronics. Besides, the LEDs in this design were powered through a PWM output rather than the 3.3 V power supply of RPi to enable adjustable lighting control through software to optimize the capture of texture details. The protective shell also protects fragile components (silicon layer and LEDs) and provides a platform to position and secure the camera (Arducam 1/4 inch 5 MP sensor mini). Both the rigid frame and protective shell of the sensor module were printed with a Stereolithography (SLA) 3D Printer (Formlabs Form 3, Formlabs Inc.) using clear resin material (FLGPCL04, Formlabs Inc.).

**Computer module** includes a Raspberry Pi 4B(RPi-4B) (Raspberry Pi Foundation, UK)- as the *edge* device, an on/off rocker switch (JF0027-1, QTEATAK), and a battery HAT board (Pisugar 2 Pro, Pisugar) with a 5000 mAh UPS lithium battery pack. The switch is connected to the RPi as an external input and fixed on the handle for easy operation. The battery capacity is tested to be sufficient operating more than one hour under the worst operating condition (i.e., when all features were implemented and working). The battery pack is connected to the Hat through magnets on each other, while the Hat is secured with RPi by screws.

**Handle** of this device adopts the aforementioned modules and provides an ergonomic holding space for human users. It is initially printed as two separate parts using the FDM 3D printer (Raise 3D Pro2 Series, RAISE3D Inc.) with the PLA material (RAISE3D INC.). The handle reserves two major opening spaces, as shown in Fig. 3 (b), for easy wire connection, debugging, and status monitoring without disassembly. Each opening is covered by a transparent 2 mm acrylic plate cut by a laser machine (Fusion Pro Co2 Laser, Epilog). These acrylic coverings ensure that the inner electronics will not be contaminated by the outside environment in the operating room. Small openings on the front and bottom side of the device are, however, reserved open for USB flash drives plugging in/out, keyboard or/and mouse receiver connection, HDMI output, and power on/off the system, as demonstrated in Fig. 3 (a). Typically, a USB flash drive was used for local data downloading after the procedure, while the HDMI and keyboard or/and mouse were optional and backup for wired

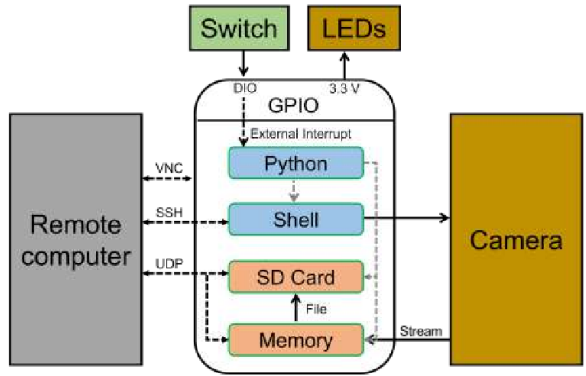

Fig. 4. Handheld edge device software architecture: the center block symbolizes the hardware and software interaction inside the RPi; Camera, LEDs, and Switch are peripherals that are essential to the handheld system; the remote computer is optional but provides remote assistance and monitoring to the operation of the device.

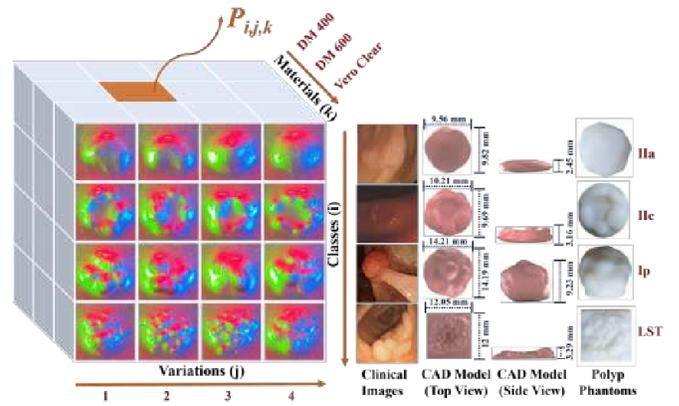

Fig. 5. A conceptual three-dimensional tensor showing the dimensions of the four types of fabricated CRC polyp phantoms that is utilized to collect data by the HySenSe

screening and control through a local computer. Although they are not covered by acrylic plates, simple flexible plastic wrap can prevent them from being exposed to environmental settings. (i.e., the blood of the tissue).

### B. Software Architecture

Fig. 4 demonstrates the software architecture for the proposed smart handheld edge device. This architecture overall realizes three statuses. After the device is powered on, the microprocessor in RPi will persistently monitor the external interrupt of a specific GPIO port connected with the rocker switch. This interrupt was configured to detect both the rising and falling edge of the GPIO signal and switch the status accordingly. The *working* (or Visual Data Collection ON) status will be activated when the user toggles the rocker switch to ON and the rising edge of the signal is captured by the interrupt detection. This status signified the beginning of the video streaming and recording (saved temporally in the local memory of RPi), which was running as a thread in the Raspberry Pi OS (Bulleye, August 2019). The video stream can be accessed and simultaneously forwarded to a remote PC through User Datagram Protocol (UDP). The last status is *Polyp Interaction* which can be regarded as a special state during the working status we described above, except that the gel layer started to interact with the objects. Python is primarily used as the programming environment to communicate with the dedicated GPIO and detect the external interrupt through APIs from *RPi.GPIO*. ON/OFF commands for the camera, however, are implemented with Shell commands which are accessed through *os* APIs in Python. Notably, Secure Shell Protocol (SSH) and Virtual Network Computing (VNC) provide remote control/monitoring access from computers under the same network.

### C. Polyp Phantoms

To thoroughly evaluate the performance and sensitivity of the proposed handheld device, and particularly HySenSe, we designed and fabricated 48 different types of CRC polyp phantoms (4 CRC polyp types, 4 geometric variations of each type, 3 different materials) mimicking the topology, size, and texture of CRC polyps based on realistic clinical cases using the Digital Anatomy Printer (J750, Stratasys, Ltd) [23]. These polyps replicate the pedunculated (type Ip), laterally spreading tumor (type LST), superficial elevate (type IIa), and depressed (type IIc) that represent the Paris classification [21]. As the tensor shown in Fig.5 demonstrates, each unique polyp $P_{i,j,k}$ is characterized by its aforementioned Paris classification $i$ [21], the geometric textural variation $j$, and the material $k$, which represents varying stiffness characteristics [24]. Of note, we used materials with varying hardnesses as 00 45-60, A 30-40, and D 83-86 for Tissue Matrix/Agilus DM 400, the mixture of Tissue Matrix and Agilus 30 Clear, and Vero Pure White, respectively [25]. More details about the fabrication can be found in [16].

### D. Texture and Stiffness Classification AI Models

In this paper, we propose a dual-stage machine learning model that maximizes the performance of polyp detection and stiffness classification by separating the aforementioned two classification problems. The proposed approach will first input the textural occlusion-free illumination-resilient information captured by the proposed handheld edge device into a dilated Convolutional Neural Network (CNN) to detect the type of a CRC polyp, as the first stage of classification. Next, in the second stage, based on the detected type, it will use a t-distributed stochastic neighbor embedding (t-SNE) [26] to estimate its corresponding stiffness. We have recently evaluated [27] the performance of a fine-tuned dilated CNN on a limited textural data set and reported its superior performance when compared with existing counterparts for polyp classification. Also, the t-SNE approach is a statistical method for nonlinear dimensionality reduction, commonly used for visualizing high-dimensional data. t-SNE has shown success in various medical applications such as EEG-based epileptic seizure detection and chest X-ray analysis of lung cancer [28], [29]. Of note, due to the highly nonlinear deformation behavior of an elastic object like the HySenSe gel layer and CRC polyps, t-SNE is able to capture such complex and high-dimensional deformation.

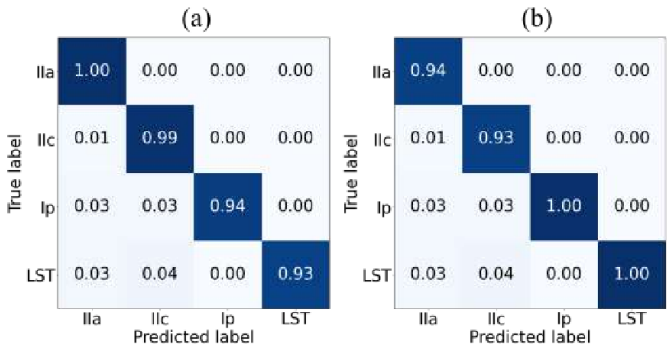

Fig. 6. Normalized confusion matrices for our dilated CNN's texture classification across four folds for (a) class-wise sensitivity and (b) class-wise precision.

### E. Experimental Setup and Procedure

To correlate the applied interaction force and the high-quality HySenSe visual outputs, and obtain data for the training of the real-time polyp type detection and hardness classification, we used an experimental setup and procedure described in [16]. This setup consisted of the sensor module of the handheld device, the fabricated polyp phantoms, a precise force gauge, and a linear stage to collect the mentioned data. To conduct experiments, first, one of the fabricated polyps was connected to the force gauge. Then, the sensor module was fixated on an optical table while the linear stage was pushing the CRC polyp phantom on the HySenSe surface. During this procedure, the recorded textural images and corresponding interaction forces from 0-0.8 N were recorded. These steps were then repeated for all 48 CRC phantom polyps. The recorded textural images and the interaction forces were used to train and evaluate the proposed dual-stage type and stiffness classification procedure.

The dataset of 48 unique polyps obtained using these experiments was then augmented by 6 times, resulting in a total dataset size of 288 samples. The pre-processing of the data consisted of first cropping the output textural frame of HySenSe to only include the polyp area of interest and resizing the images to 224×224 pixels, as required by our dilated CNN. Next, the dataset was augmented using geometric transformations such as horizontal flips, random rotations in the range of $-90^o$ to $90^o$, and a combination of flips and rotations. Lastly, a Gaussian Blur augmentation was used with a blur factor $\sigma$ chosen randomly in the range of 32 to 64 for each image.

## III. RESULTS

The dilated CNN was trained for 15 epochs using a stratified k-folds approach using four folds across 80% of the data (i.e., 216 samples). Of note, a pre-trained network was used and the model's weights were fine-tuned to best complement our dataset. The AdaBound optimizer [30] was used for the model training, and an initial learning rate of 0.001 and a final learning rate of 0.01 were found to produce the best results for our data. Following the k-folds training process, the best-performing weights of each of the four folds were used to evaluate the model. The remaining 20% of the data (i.e., 72 samples) not used for training, evenly distributed across each Paris type, was reserved for model evaluation.

It is clinically important to consider the model's performance with respect to each class pair by analyzing class-wise sensitivity $C_S^{i,j}$ and class-wise precision $C_P^{i,j}$. $C_S^{i,j}$ evaluates the ability of the model to correctly label the polyp class and $C_P^{i,j}$ evaluates the correctness of the prediction. Normalized class-wise sensitivity and class-wise precision confusion matrices were constructed (Fig. 6) by combining the model's predictions on the evaluation data (i.e., 72 samples) across the best-performing weights for each of the four training folds.

In addition, performance metrics such as accuracy ($E_{acc}$), recall ($E_{rec}$), specificity ($E_{spec}$), and precision ($E_{prec}$) were used to further evaluate the classification capability of our model. These metrics are significant in reliable CRC polyp classification by providing a thorough evaluation that goes beyond just accuracy. $E_{acc}$ represents the model's ability to make correct predictions across all the texture classes. $E_{rec}$ and $E_{spec}$ measure the model's true positive and true negative predictions, respectively. Of note, recall ($E_{rec}$) is the most widely utilized metric to evaluate the performance of deep-learning models (e.g., our dilated CNN) for CRC polyp classification tasks [31]. Finally, $E_{prec}$ evaluates the model's confidence when predicting a particular class.

To identify the stiffness and stage of the known type of a polyp classified by the previous model, the t-SNE algorithm was utilized. In this experiment, we chose 24 of the 48 polyps to demonstrate t-SNE results on all four polyp classes across two geometric variations from each class and all three stiffnesses. Each plot visualizes one polyp and its respective counterparts across the three materials. Images were resized to 224×224 pixels and converted to grayscale. PCA was chosen as the initialization of embedding, and a random seed was chosen to ensure reproducible results. For t-SNE algorithm hyperparameters such as perplexity, learning rate, and the number of iterations are crucial, tunable parameters. As suggested by [26], the perplexity parameter was optimized from a range of 5-50, and it was found that a value of 5 worked best for our data. The learning rates were optimized for each run, and the model was run for 5000 iterations to ensure a stable configuration was reached for each set of data. Fig. 7 visualizes the results of the algorithm for applied interaction forces between the HySenSe and each polyp in a range of 0.2-0.8 N. Images that belonged to the same material were clustered and highlighted using the colors represented in the legend shown.

## IV. DISCUSSION

Our dilated CNN used for textural classification consistently performed strongly in classifying each of the four classes with class-wise sensitivity $\geq 0.93$ (Fig. 6a) and class-wise precision $\geq 0.93$ (Fig. 6b). The high class-wise sensitivity and precision values of $C_S^{IIa,IIa} = 1.00$, $C_S^{LST,LST} = 0.93$, $C_P^{IIa,IIa} = 0.94$, $C_P^{LST,LST} = 1.00$ exemplify the model's capability in distinguishing challenging cases. Of note, IIa and LST-type textures are regarded to be "flat polyps", and therefore are the most difficult to diagnose during screening [32].

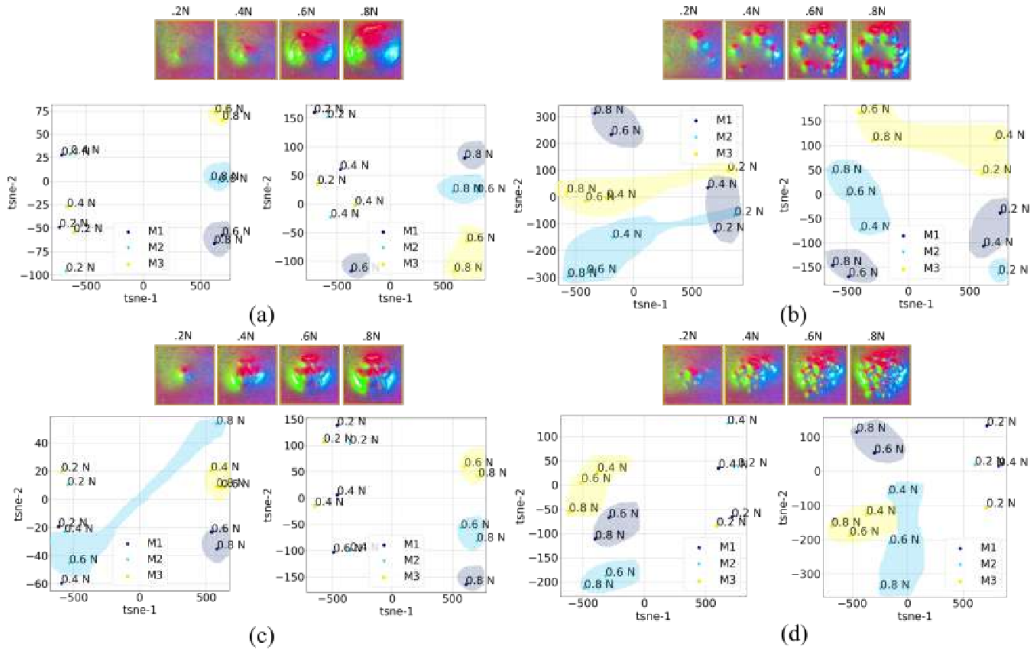

Fig. 7. The classification of polyp stiffness (M1, M2, M3) using the t-SNE algorithm for two variations of each CRC polyp type along with HySenSe outputs of polyp types across different force thresholds (0.2 N - 0.8 N). (a) IIa (b) IIc, (c) Ip, (d) LST.

Furthermore, type IIa and IIc are known as non-polypoid colorectal tumors, and these are often missed by inexperienced colonoscopists [23]. The IIc types are slightly depressed polyps, which have an increased risk of invading the submucosa [23]. The results in Fig. 6a show that the class-wise sensitivity for IIa and IIc are $C_S^{IIa,IIa} = 1.00$ and $C_S^{IIc,IIc} = 0.99$, thereby revealing the strong potential to aid colonoscopists in identifying these polyps types and thus decrease the risk submucosal invasion. The model achieves a high average accuracy of 98.26% across the four folds, which supports its potential to correctly label the Paris textural classes. Similarly, the model exhibits an average recall and specificity of 96.53% and 98.84%, respectively, demonstrating its ability to reliably identify true positives and true negatives. This is significant in effectively distinguishing the polyp types that are at a high risk of invading the submucosa. Finally, an average precision of 96.73% was recorded.

To investigate the polyp's cancerous nature, its modulus of elasticity and stiffness needs to be also analyzed. Fig. 7 visualizes the stiffness classification results on all four types of polyps made from the three different materials (M1, M2, M3) across four images at different thresholds (0.2, 0.4, 0.6, and 0.8 N). Interestingly, the HySenSe outputs from the polyps with the M3 material are isolated from the M1 and M2 material clusters, and this is important as polyps with high modulus of elasticity are associated with a more cancerous potential [6]. As seen through the plots, images across different stiffness levels are generally distinguishable over 0.6 N by forming distinct clusters. The interaction forces between the HySenSe and polyps should be under 13.5 N to be safe and not cause damage to the colon tissue [22]. The interaction force of 0.6 N to effectively distinguish polyp stiffness, therefore, is well below the 13.5 N threshold demonstrating the capability of the proposed dual-stage classification algorithm for simultaneous detection of type and stage of CRC polyps.

TABLE I
DIMENSIONS OF REALISTIC POLYP PHANTOMS

| Pressure | Loading Coefficients | | | Unloading Coefficients | | Hysteresis | |
|---|---|---|---|---|---|---|---|
| | a | h | k | a | b | Area | Relative |
| $P_0 = 0$ | 0.09 | 1.61 | 0.13 | 0.59 | -1.25 | 2.1 | 1.00 |
| $P_1 = 2.17 kPa$ | 0.37 | 1.89 | 0.26 | 1.04 | -2.15 | 0.33 | 0.16 |
| $P_2 = 2.74 kPa$ | 0.57 | 2.78 | 0.48 | 1.54 | -4.59 | 0.38 | 0.18 |
| $P_3 = 3.21 kPa$ | 0.32 | 2.97 | 0.57 | 1.10 | -3.33 | 0.39 | 0.19 |

## V. CONCLUSION AND FUTURE WORK

In this study, to collectively address the existing challenges in performing an on-site CRC polyp diagnosis and pathology, we proposed a unique framework that includes a smart handheld edge device enabling clinicians to obtain high-resolution textural images using a unique occlusion-free, illumination-resilient tactile sensing device. We also proposed a dual-stage machine learning algorithm that can reliably and sensitively ($> 97\%$) identify the type and stage of CRC polyps by classifying their texture and stiffness, respectively. The performance of the proposed framework was successfully evaluated on unique CRC polyp phantoms and utilizing appropriate statistical metrics. In the future, we will plan to perform an online CRC polyp classification and also evaluate the performance of the proposed edge framework in a real clinical setting inside the operating room.